\pdfoutput=1

\documentclass[11pt]{article}

\usepackage{acl}

\usepackage{times}
\usepackage{latexsym}

\usepackage[T1]{fontenc}

\usepackage[utf8]{inputenc}

\usepackage{microtype}

\usepackage{inconsolata}

\usepackage{graphicx}

\usepackage{booktabs}
\usepackage{multirow}
\usepackage{multicol}

\usepackage{tcolorbox}
\usepackage{hyperref}
\usepackage{textcomp}

\usepackage{xcolor} 

\newcommand{\trace}{\textsc{Trace}} %

\title{\trace: Real-Time Multimodal Common Ground Tracking in Situated Collaborative Dialogues}

\author{
 \textbf{Hannah VanderHoeven\textsuperscript{1}},
 \textbf{Brady Bhalla\textsuperscript{2*}},
 \textbf{Ibrahim Khebour\textsuperscript{1}}, 
 \textbf{Austin Youngren\textsuperscript{1}},
\\
 \textbf{Videep Venkatesha\textsuperscript{1}},
 \textbf{Mariah Bradford\textsuperscript{1}},
 \textbf{Jack Fitzgerald\textsuperscript{1}},
 \textbf{Carlos Mabrey\textsuperscript{1}},
 \textbf{Jingxuan Tu\textsuperscript{3}},
 \\
 \textbf{Yifan Zhu\textsuperscript{3}},
 \textbf{Kenneth Lai\textsuperscript{3}},
 \textbf{Changsoo Jung\textsuperscript{1}},
 \textbf{James Pustejovsky\textsuperscript{3}} \and
 \\
 \textbf{Nikhil Krishnaswamy\textsuperscript{1}}
\\
\\
 \textsuperscript{1}Colorado State University, Fort Collins, CO USA;\\
 \textsuperscript{2}California Inst. of Technology, Pasadena, CA USA;
 \textsuperscript{3}Brandeis University, Waltham, MA USA
\\
 \small{
   \textbf{Correspondence:} \href{mailto:hannah.vanderhoeven@colostate.edu}{hannah.vanderhoeven@colostate.edu}, \href{mailto:nkrishna@colostate.edu}{nkrishna@colostate.edu}
 }
}

\begin{document}
\maketitle
\def\thefootnote{*}\footnotetext{This work performed under a CalTech Summer Undergraduate Research Fellowship (SURF) program at Colorado State University.}\def\thefootnote{\arabic{footnote}}

\begin{abstract}

We present \trace, a novel system for live {\it common ground} tracking in situated collaborative tasks. With a focus on fast, real-time performance, \trace~tracks the speech, actions, gestures, and visual attention of participants, uses these multimodal inputs to determine the set of task-relevant propositions that have been raised as the dialogue progresses, and tracks the group's epistemic position and beliefs toward them as the task unfolds.  Amid increased interest in AI systems that can mediate collaborations, \trace~represents an important step forward for agents that can engage with multiparty, multimodal discourse.

\end{abstract}

\section{Introduction}
\label{sec:intro}
\vspace*{-2mm}

When engaging in a shared task, collaborators continually exchange information about goals, obstacles, and next steps, thereby building a shared understanding of the problem, or a ``common ground''~\cite{clark1991grounding}. 
In situations involving hybrid human-AI teams, although there is an increasing desire for AIs that act as collaborators with humans, modern AI systems struggle to account for such mental states in their human interlocutors~\cite{sap2022neural,ullman2023large} that might expose shared or conflicting beliefs, and thus predict and explain in-context behavior~\cite{premack1978does}. Additionally, in realistic scenarios such as collaborative problem solving~\cite{nelson2013collaborative}, beliefs are communicated not just through language, but through multimodal signals including gestures, tone of voice, and interaction with the physical environment~\cite{vanderhoeven2024multimodal}. Since one of the critical capabilities that makes human-human collaboration so successful is the human ability to interpret multiple coordinated modalities in real-time, collaborative AIs would need to likewise replicate this ability in live real-time settings, but this remains extraordinarily difficult for machines.

\begin{figure}
    \centering
    \includegraphics[width=.48\textwidth]{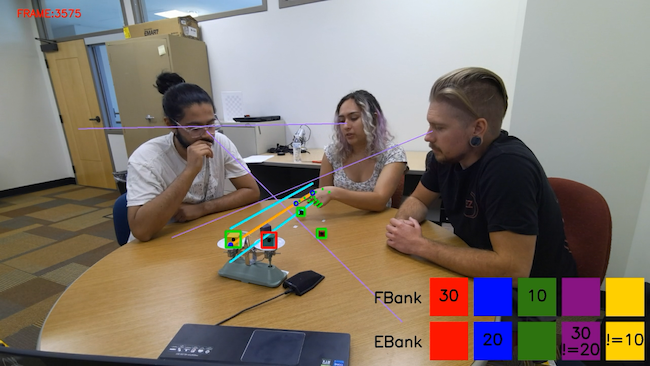}
\vspace*{-4mm}
    \caption{Three participants performing the Weights Task with overlay showing detected deixis, objects, and gaze directions, as well as banks of {\it evidence} ({\sc EBank}) and agreed-upon {\it facts} ({\sc FBank}) regarding the weights of each differently-colored block.}
    \label{fig:teaser}
\vspace*{-4mm}
\end{figure}


Our system, \trace~(Transparency in Collaborative Exchanges) addresses this problem with the following novel and unique contributions in a single system:
\begin{itemize}
\vspace*{-1mm}
    \item Real-time tracking of participant speech, actions, gesture, and gaze when engaging in a shared task;
\vspace*{-1mm}
    \item On-the-fly interpretation and integration of multimodal signals to provide a complete scene representation for inference;
\vspace*{-1mm}
    \item Simultaneous detection of asserted propositional content and epistemic positioning to infer task-relevant information for which evidence has been raised, or which the group has agreed is factual;
\vspace*{-1mm}
    \item A modular, extensible architecture adaptable to new tasks and scenarios.
\vspace*{-1mm}
\end{itemize}

We demonstrate \trace~on the task of tracking the {\it common ground} that emerges within triads performing a situated collaborative task called the Weights Task~\cite{khebour2024text} (Fig.~\ref{fig:teaser}). Importantly, our system jointly operationalizes methods previously evaluated in isolation~\cite{khebour-etal-2024-common-ground,vanderhoeven2024point,venkatesha2024propositional}, and we do this in \textit{real-time} while balancing speed and performance. To our knowledge, no previous system has attempted this. \trace~can be adapted to similar situated collaborative tasks with sufficient data, making it useful for real-time analysis of collaborative problem solving and multimodal communication. We also assess the level of error introduced into multiple features by different levels of live automated processing when compared to manually-annotated ground truth. \trace~represents an important advance for for AI systems that can model group collaboration in real-time situated contexts. A video demonstration showcasing multiple aspects of a collaborative interaction is available \href{https://youtu.be/BAuWCMbC6Ls}{here}. Installable code and setup instructions may be found at \url{https://github.com/csu-signal/TRACE/releases/tag/naacl-demo}, available at present under the MIT license.

\vspace*{-2mm}
\section{Related Work}
\label{sec:related}
\vspace*{-2mm}

Dialogue state tracking (DST) aims to update the representations of a speaker's (user’s) needs at each turn in the dialogue, taking into account past dialogue moves and history \cite{budzianowski-etal-2018-multiwoz,liao-etal-2021-dialogue,jacqmin-etal-2022-follow}.  Dialogue studies provide a technical definition of a ``common ground'' as a set of shared beliefs among participants in an interaction~\cite{grice1975logic,clark1991grounding,traum1994computational,stalnaker2002common,asher2003common,traum2003information,hadley2022review}. This attribution of mental states to one's interlocutors is central to {\it Theory of Mind}~\cite{premack1978does}. Such internal states may be communicated not just through language, but nonverbal behavior as well~\cite{hall2019nonverbal}. Understanding nonverbal behavior in multimodal communication has been of longstanding interest in psychology and HCI \cite{kendon1997gesture,kendon2004gesture,mcneill2005gesture,beilock2010gesture}, and has recently found increasing relevance to AI systems \cite{sigurdsson2016hollywood,gu2018ava,li2020ava}.

Our work is similar in spirit to the Dialogue State Tracking Challenge (DSTC; \citet{williams2016dialog}). While both are consistent with \citet{clark1996using}'s notion of common ground and may involve a live evaluation, our work is novel in that we address the content of the common ground directly rather than proxies such as goal, and interpret multimodal signals in a situated collaborative task context. Similar work that involves situated interaction includes grounding of action descriptions~\cite{beinborn2018multimodal}, and previous work using interactive virtual avatars~\cite{krishnaswamy2017communicating,pustejovsky2017creating,krishnaswamy2020diana} where a common ground can be constructed {\it post hoc}~\cite{krishnaswamy2020formal}.

\citet{khebour-etal-2024-common-ground}, introduced a novel task of {\it common ground tracking} (CGT) that automatically identifies the set of shared beliefs and “questions under discussion” (QUDs) of a group with a shared situated task and goal, using multimodal signals to both extract the propositional content being expressed by task participants \cite{venkatesha2024propositional}, and their epistemic positionings toward them, to mark which are accepted as facts by the group vs. merely evidenced. \citet{vanderhoeven2024multimodal} laid out the different modalities that may be used to give an AI system enough information to adequately interpret a collaborative dialogue. \trace~operationalizes and integrates the aforementioned works in real time.

\vspace*{-2mm}
\section{System Description}
\label{sec:desc}
\vspace*{-2mm}

\begin{figure}
    \centering
    \includegraphics[width=0.6\textwidth]{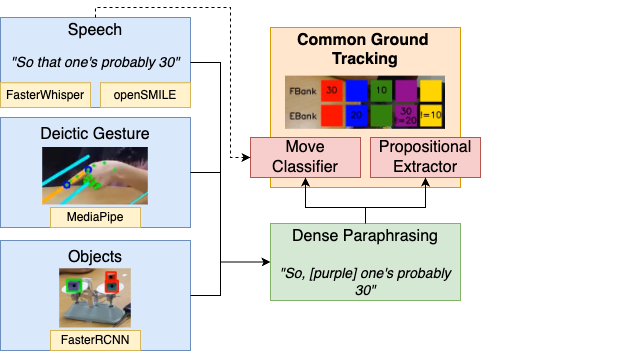}
\vspace*{-4mm}
    \caption{High-level schematic of information flow in real-time multimodal common ground tracking. We combine signals from speech, gesture, and objects in the environment to determine the task-relevant content being discussed, and the epistemic positioning expressed in each utterance. Logical closure rules unify these outputs into the set of common QUDs ({\sc QBank}---not displayed for space reasons), pieces of evidence ({\sc EBank}), and facts ({\sc FBank}).}
    \label{fig:diagram}
\vspace*{-4mm}
\end{figure}

\trace~is a modular system that combines features from speech, acoustic, RGB, and depth channels to interpret task participants' linguistic and nonverbal behavior to model their common task-relevant beliefs. Descriptions of the individual modules and their relations to previous research are given in Sec.~\ref{ssec:modules}, and Fig.~\ref{fig:diagram} shows how they interact.  All feature modules specify an output {\it interface} or a class representing the data type a module outputs. Modules also specify zero or more input interfaces, which they require in order to calculate the output. For example, the Propositional Extraction module requires only text input while the Dense Paraphrasing module requires text, gesture, and object inputs (Fig.~\ref{fig:diagram}). \trace~enables modules to set their input interfaces as dependencies, and the contents of the required output interface will be automatically passed into the dependent input interface. Thus, the entire system, or any such system built with \trace~can be structured as a directed graph with features as vertices and edges connecting a module and all of its dependencies. This framework allows for swapping in and out different multimodal processing modules to create variants of the system.

\trace~is demonstrated on the Weights Task~\cite{khebour2024text}, a situated collaborative task where triads work together to determine the weights of five differently-colored blocks using a balance scale. The block weights follow the pattern of the Fibonacci sequence in increments of 10 grams. The correct weight assignments by color are: 10g (red), 10g (blue), 20g (green), 30g (purple), and 50g (yellow). Beliefs in the Weights Task constitute evidence for or against weight assignments for blocks, or agreement upon the weight of a given block, as discussed in \citet{khebour-etal-2024-common-ground}. Fig.~\ref{fig:teaser} shows the physical task space, with blocks and the balance scale on a table with 3 participants seated around it. The task is recorded using an Azure Kinect RGBD camera, and either a single MXL AC-404 ProCon microphone or 3 individual lavalier or headset mics---one for each participant \cite{bradford2022deep}. The system as presented in the demonstration video runs on an Alienware Aurora R12 tower with an NVIDIA RTX 3090 with 24GB of VRAM but can run on systems as small as a laptop with an RTX 3070 Ti (8GB VRAM). See Appendix~\ref{app:profiling} for further details.

\vspace*{-2mm}
\subsection{Modules}
\label{ssec:modules}
\vspace*{-1mm}

Here we describe the individual modules used by \trace~for multimodal processing. Our choices of processing techniques were motivated by the need to simultaneously optimize for both performance and the speed necessary to run in real time while remaining within the aforementioned hardware limits when running all modules simultaneously. Thus, we combine older and newer techniques to provide sufficient performance while running quickly enough for real-time processing. The technical details of each are in the referenced papers. Details such as hyperparameters or minor modifications we made to the original models are deferred to Appendix~\ref{app:modules}.

\vspace*{-2mm}
\paragraph{Automatic Speech Recognition}
For automatic speech recognition, we use the FasterWhisper variant of Whisper~\cite{radford2023robust}. Acoustic and prosodic features of utterances are extracted using openSMILE \cite{eyben2010opensmile}.

\vspace*{-2mm}
\paragraph{Object Detection}
Detection of the blocks in the scene uses a FasterRCNN ResNet-50-FPN model \cite{lin2017feature} trained over block bounding box annotations from the original Weights Task Dataset (WTD; \citet{khebour2024text}).

\vspace*{-2mm}
\paragraph{Deictic Gesture and Gaze Detection}
We use the 3-stage gesture recognition method from \citet{vanderhoeven2023robust} that operationalizes the gesture semantics of \citet{kendon1997gesture} to detect the ``stroke'' or semantically-important phase of a gesture; e.g., for deictic gesture, this is the extension of a digit. We then use \citet{vanderhoeven2024point}'s method to calculate a ``pointing frustum''~\cite{kranstedt2006deixis} from the extended digit into 3D space and intersect it with detected objects to determine what the targets of deixis are. A similar method is used to infer gaze direction from the direction of participants' heads (see Appendix~\ref{app:modules}).

\vspace*{-2mm}
\paragraph{Multimodal Dense Paraphrasing (MMDP)}
In situated dialogue, objects are often referenced with demonstratives (``this,'' ``that one,'' etc.). Fully interpreting these demonstratives requires recourse to one or more non-linguistic modality. We follow a {\it multimodal dense paraphrasing} (MMDP) procedure \cite{tu-etal-2023-dense,tu2024dense}, which uses additional context to merge multimodal channels into enriched LLM prompts that query the state of the common ground. \trace~uses MMDP to build a list of potential referents from the objects selected by deixis, and then takes demonstratives in utterances that overlap with the deictic gesture and replaces them with the names of the objects, depending on the objects in the list (ordered by distance from the pointing digit) and the grammatical number of the demonstrative pronoun. Table~\ref{tab:dps} provides examples.

\begin{table}
    \resizebox{.49\textwidth}{!}{
    \begin{tabular}{lll}
\toprule
\textbf{Blocks} & \textbf{Utterance} & \textbf{Dense paraphrase} \\
\midrule
\multirow{1}{*}{purple} & So, \textit{that}'s more than 20 & So, \textit{[purple block]}'s more than 20. \\
\multirow{2}{*}{red, green} & So \textit{that}’s a 10 and & So \textit{[red block]}’s a 10 and \\
& \textit{that}’s a 20 right there? & \textit{[green block]}’s a 20 right there? \\
\multirow{1}{*}{green, purple} & So, \textit{these} are 50 on here? & So, \textit{[green block, purple block]}  \\
& & are 50 on here? \\
\bottomrule
	\end{tabular}}
\vspace*{-2mm}
	\caption{\label{tab:dps}Utterances, retrieved blocks, and corresponding (ground truth) MMDPs.}
\vspace*{-4mm}
\end{table}

\vspace*{-2mm}
\paragraph{Common Ground Tracking (CGT)}
CGT follows \citet{khebour-etal-2024-common-ground}'s method, combining an epistemic ``move'' classifier, a propositional extractor, and a set of logical closure rules to enforce consistency over the facts, evidence, and questions under discussion within the group's common ground. Utterances are classified as expressing an epistemic $STATEMENT$ of evidence toward the currently or most-recently expressed proposition, $ACCEPT$ance of previously-surfaced evidence as fact, $DOUBT$ of evidence or a fact, or none of the above. These classifications are performed on the basis of the MMDPed text of the transcribed utterance encoded through BERT~\cite{devlin-etal-2019-bert}, and the acoustic/prosodic features extracted with openSMILE, and does not include other features like Gesture-AMR (GAMR; \citet{brutti-etal-2022-abstract}) or collaborative problem solving facets~\cite{sun2020towards}, which require manual annotations or an auxiliary model~\cite{bradford2023automatic}. 

Propositions are extracted from the text of the dense-paraphrased utterance, and take the form of relations between blocks or between blocks and weight values (e.g., $red = 10$ or $red = blue$). Here we use the cross-encoder method from \citet{venkatesha2024propositional}, who report improved performance over the cosine similarity method used in \citet{khebour-etal-2024-common-ground}. Further technical specifications are given in Appendix~\ref{app:modules}.

Logical closure rules consistent with those in \citet{khebour-etal-2024-common-ground} unify the extracted propositions and epistemic moves into the contents of the common ground. $STATEMENT(p)$ raises evidence consistent with $p$ to {\sc EBank}. $ACCEPT(p)$ raises $p$ (if in {\sc EBank}) to {\sc FBank}. $DOUBT(p)$ lowers $p$ (if in {\sc FBank}) to {\sc EBank}. For $ACCEPT$s and $DOUBT$s, we assume $p$ to be the most-recently stated proposition if no proposition is extractable from the utterance (e.g., utterances like ``yeah'' or ``wait, I don't think so'').

\vspace*{-2mm}
\section{Evaluation}
\label{sec:eval}
\vspace*{-2mm}

We evaluate over Groups 1, 2, 4, and 5 of the Weights Task Dataset (WTD). These groups have been fully manually annotated with ground truth labels for all speech transcriptions, gestures, block locations, epistemic ``move'' labels, and expressed propositions. We use move and proposition models that exclude the relevant test group from the training data. We also present an evaluation of the demonstration video (linked in Sec.~\ref{sec:intro}) where the models used were trained over the entire WTD.

\begin{figure}
    \centering
\vspace*{-4mm}
    \includegraphics[width=0.48\textwidth]{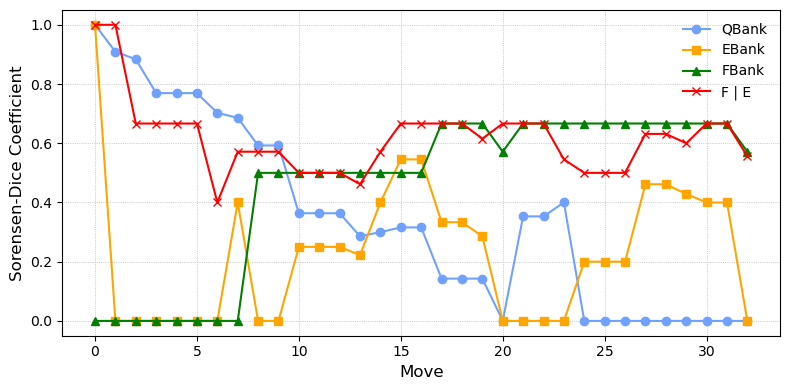}
\vspace*{-6mm}
    \caption{DSC of each common ground bank vs. moves in the demo video dialogue. A value may be zero if there is no intersection between the predicted and ground truth sets, but also if the union of the ground truth and predicted sets for that bank is empty, resulting in zero denominator. We treat this case as no similarity.}
    \label{fig:dsc-demo}
\vspace*{-4mm}
\end{figure}

Following \citet{khebour-etal-2024-common-ground}, our primary metric is S\o rensen-Dice Coefficient (DSC; \citet{dice1945measures,sorensen1948method}).  This can be computed against each utterance (as in Fig.~\ref{fig:dsc-demo}), or averaged over a dialogue (as in Table~\ref{tab:live-vs-cgt}), and indicates the match between the set of propositions extracted by \trace~using all the component models, and the set of propositions in the ground truth, while also normalizing for the size of the two sets.

We compare \trace's live performance to \textit{post hoc} results from \citet{khebour-etal-2024-common-ground}, who considered only utterances that were annotated as expressing some epistemic position, and used human-annotated dense paraphrases and gesture annotations using GAMR~\cite{brutti-etal-2022-abstract}. We present DSC over all three common ground banks, as well as over the union of {\sc FBank} and {\sc EBank}, which approximates the quality of propositional extraction independent of epistemic move classification (because misclassified moves may raise a proposition $p$ to the wrong bank). Due to the challenge of real-time processing, our reported numbers are often lower, though we do find a few cases where we match or slightly exceed previous results, such as extracting QUDs in Group 5. Generally, live processing does fairly well at tracking the set of QUDs over time but struggles to assign facts and evidence to the right level. This was also a challenge noted in the original \citet{khebour-etal-2024-common-ground} results.

As such, we also compare to results reported in \citet{tu2024dense}, who focus on using multimodal dense paraphrasing to identify common ground in the aftermath of human-labeled $ACCEPT$ moves, and hence only report results on {\sc FBank}. Thus, their results can be directly compared to the union of facts and evidence ({\sc F $\cup$ E}) in the live condition, as they implicitly assume the contents of other preceding utterances accumulate evidence which is then moved to fact status upon the occurrence of a human-labeled $ACCEPT$.

This represents comparisons to all previously-reported SOTA on this task and data, however our numbers represent real-time automated processing of {\it all features} considering all utterances (unlike \citet{khebour-etal-2024-common-ground}), meaning that we are simultaneously \textit{detecting and classifying} epistemic positioning, \textit{and} considering all banks of the common ground (unlike \citet{tu2024dense}). Table~\ref{tab:live-vs-cgt} shows the results and comparisons.

\begin{table}[h!]
    \begin{tabular}{lllll}
\toprule
 & \small{\bf Group 1} & \small{\bf Group 2} & \small{\bf Group 4} & \small{\bf Group 5} \\
\midrule
 \multicolumn{5}{c}{\small\trace} \\
\midrule
\small{\bf QBank} & \small 0.349 & \small 0.656 & \small 0.741 & \small 0.546 \\
\small{\bf EBank} & \small 0.063 & \small 0.135 & \small 0.231 & \small 0.214 \\
\small{\bf FBank} & \small 0.000 & \small 0.205 & \small 0.000 & \small 0.000 \\
\small{\bf F $\cup$ E} & \small 0.246 & \small 0.377 & \small 0.231 & \small 0.464 \\
\midrule
  \multicolumn{5}{c}{\small\citet{khebour-etal-2024-common-ground}} \\
\midrule
\small{\bf QBank} & \small 0.767 & \small 0.911 & \small 0.817 & \small 0.514 \\
\small{\bf EBank} & \small 0.344 & \small 0.713 & \small 0.812 & \small 0.335 \\
\small{\bf FBank} & \small 0.000 & \small 0.528 & \small 0.045 & \small 0.165 \\
\small{\bf F $\cup$ E} & \small 1.000 & \small 0.922 & \small 0.832 & \small 0.959 \\
\midrule
  \multicolumn{5}{c}{\small\citet{tu2024dense}} \\
\midrule
\small{\bf F (TA)} & \small 0.883 & \small 0.580 & \small 0.450 & \small 0.652 \\
\midrule
  \multicolumn{5}{c}{\small{\sc GPT-4o} (from \citet{tu2024dense})} \\
\midrule
\small{\bf F (TA)} & \small 0.841 & \small 0.331 & \small 0.321 & \small 0.478 \\
\bottomrule
	\end{tabular}
\vspace*{-2mm}
	\caption{\label{tab:live-vs-cgt}Comparison of \trace~live tracking performance to {\it post hoc} results from other methods. {\bf F (TA)} represents the contents of {\sc FBank} when only ``True Accepts'' are considered, as in \citet{tu2024dense}, which is equivalent to {\bf F $\cup$ E} in the real-time condition.} 
\vspace*{-4mm}
\end{table}

Like previous results, we see significant variance across groups, indicating the intrinsic challenge of the common ground tracking task. {\sc Trace} overpredicts $STATEMENT$s and underpredicts $ACCEPT$s in Groups 4 and 5, just like \citet{khebour-etal-2024-common-ground}. This leads to propositions correctly being surfaced as evidence but never raised to facts according to the model. We also find that \trace~approaches or outperforms GPT-4o (as reported in \citet{tu2024dense}) on fact retrieval in Groups 2 and 5 given the assumption of true $ACCEPT$ classification.

Fig.~\ref{fig:dsc-demo} shows an evaluation over the demonstration video, showing detected common ground vs. the annotated ground truth over time per utterance. This shows a typical pattern in the evolution of common ground as the task unfolds, where the group begins with a full set of QUDs, over time evidence is surfaced (shown as peaks in {\sc EBank}), and certain correct facts are agreed upon over time.

Sequences in the demonstration video can also be explained by individual module performance. The utterance ``okay, I guess this one's 10'' is correctly paraphrased as ``okay, I guess [red] one's 10'' using gesture and object signals, and the correct proposition $red=10$ is extracted. However, the move classifier predicts that the utterance is an $ACCEPT$ (correct label is $STATEMENT$). Since $red=10$ is not already in {\sc EBank} here, $red=10$ is not raised to {\sc FBank}. Later, the utterance sequence ``so purple is 30 and blue is 10?'' (with pointing) and ``yeah, that should be 40 right there'' raise both $purple=30$ and $blue=10$ from {\sc EBank} to {\sc FBank} simultaneously.

\vspace*{-2mm}
\subsection{Substitution Study}
\label{ssec:abl}
\vspace*{-1mm}

Because we have the ground truth annotations for speech transcriptions, gestures, and block locations, we perform a substitution study following \citet{cohen1988evaluation} to quantify of the level of error introduced into the final output by automated processing of these features. This study is conducted by evaluating over a video as if live, except instead of passing the model outputs a given feature into the CGT pipeline, we pass the ground truth values. This allows us to evaluate the impact of each module's actual performance on the whole pipeline when compared to a hypothetical scenario where that module performs perfectly. Because of the nature of dependencies between features (see Sec.~\ref{sec:desc}), fully removing these features would prevent the system for operating entirely, and so a standard ablation study is not realistic, hence our framing of a substitution study (see Appendix~\ref{app:substitution} for more). However, given \trace's many interlinked components, such evaluation is critical to understand where specific components can be improved.

\begin{table*}
\centering
    \resizebox{\textwidth}{!}{
    \begin{tabular}{lllllllllllllll}
\toprule
& \multicolumn{4}{c}{Ground truth utterances} && \multicolumn{4}{c}{Ground truth gestures} && \multicolumn{4}{c}{Ground truth objects} \\
\cmidrule(lr){2-5} \cmidrule(lr){7-10} \cmidrule(lr){12-15}
 & \small{\bf Group 1} & \small{\bf Group 2} & \small{\bf Group 4} & \small{\bf Group 5} &
 & \small{\bf Group 1} & \small{\bf Group 2} & \small{\bf Group 4} & \small{\bf Group 5} &
 & \small{\bf Group 1} & \small{\bf Group 2} & \small{\bf Group 4} & \small{\bf Group 5} \\
\cmidrule(lr){2-5} \cmidrule(lr){7-10} \cmidrule(lr){12-15}
\small{\bf QBank} & \small 0.423 & \small 0.498 & \small 0.714 & \small 0.549 &
& \small 0.343 & \small 0.634 & \small 0.783 & \small 0.570 &
& \small 0.351 & \small 0.657 & \small 0.762 & \small 0.554 \\
\small{\bf EBank} & \small 0.031 & \small 0.042 & \small 0.248 & \small 0.263 &
& \small 0.050 & \small 0.147 & \small 0.280 & \small 0.290 &
& \small 0.067 & \small 0.135 & \small 0.231 & \small 0.247 \\
\small{\bf FBank} & \small 0.054 & \small 0.183 & \small 0.247 & \small 0.000 &
& \small 0.053 & \small 0.202 & \small 0.000 & \small 0.000 &
& \small 0.204 & \small 0.228 & \small 0.000 & \small 0.000 \\
\small{\bf F $\cup$ E} & \small 0.383 & \small 0.324 & \small 0.419 & \small 0.555 &
& \small 0.384 & \small 0.377 & \small 0.368 & \small 0.608 &
& \small 0.220 & \small 0.405 & \small 0.255 & \small 0.508 \\
\bottomrule
	\end{tabular}
}
\vspace*{-2mm}
	\caption{\label{tab:ablation}Substitution study results over the 4 WTD test groups, where instead of automatically processing the indicated feature, the ground truth value from the annotated data is passed into the rest of the pipeline.}
\vspace*{-4mm}
\end{table*}

Table~\ref{tab:ablation} shows substitution study results over the 4 WTD test groups. When using ``ground truth utterances,'' these are passed into the automated move classifier and propositional extraction models, and MMDP is performed using the automated pointing outputs. ``Ground truth gestures'' indicates that MMDP uses ground truth pointing annotations, automatically transcribed utterances, and automatically detected blocks. Likewise, ``ground truth objects'' indicates that MMDP uses automatically transcribed utterances and automatically detected points, but ground truth object bounding boxes to ensure no missed or misclassified blocks (e.g., the object detector model often confuses the blue and purple blocks due to their similar colors).

\begin{figure}
    \centering
    \includegraphics[width=.48\textwidth]{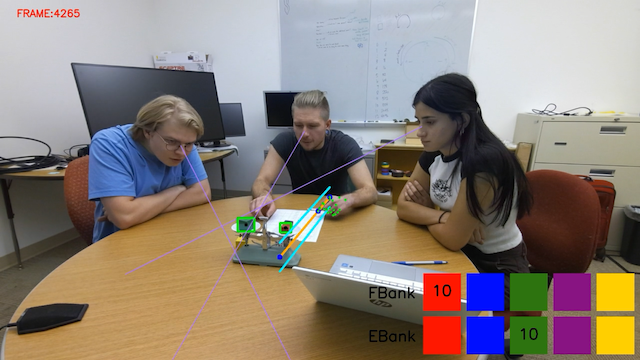}
\vspace*{-4mm}
    \caption{Still from Group 2 showing both a false positive and false negative pointing detection.}
    \label{fig:g2-still}
\vspace*{-6mm}
\end{figure}

Using veridical values for different features often significantly boosts live performance of the other modules across the board. This is most evident when using ground truth utterance transcriptions, indicating that small improvements in live ASR (e.g., correctly transcribing ``that'' instead of ``the'') would have a pronounced positive effect. Using ground truth pointing annotations is most helpful in situations like the one shown in Fig.~\ref{fig:g2-still}. Here, gesture recognition falsely detects deixis on the middle participant's left hand but misses it on the right hand. The accompanying utterance is ``now the first go through {\it it} bounced twice and actually...'' When using the ground truth pointing (annotated on the right hand, which is pointing to the green block), the MMDPed utterance is ``now the first go through {\it [green]} bounced twice and actually...'', which later helps in the correct classification of $STATEMENT(green=20)$. This shows how small improvements in an individual feature can result in substantial overall performance increase. 

Where using ground truth values adversely impacts performance, this indicates that the ground truth annotations themselves may be noisy. For instance, overlapping speech in the original Group 2 video led some utterances to be omitted from the manual transcription, meaning that ASR picked up some contentful speech that was absent in the ostensible ``ground truth''. Annotations of pointing frames are likewise also somewhat conservative.


\vspace*{-2mm}
\section{Conclusion}
\label{sec:conc}
\vspace*{-2mm}


\trace~addresses the already-challenging problem of common ground tracking in a situated collaborative task, and undertakes the added novel challenge of doing so in real time with live processing of multimodal signals. We integrate epistemic state classification \cite{khebour-etal-2024-common-ground}, propositional extraction \cite{venkatesha2024propositional}, dense paraphrasing \cite{tu-etal-2023-dense,tu2024dense}, and gesture detection \cite{vanderhoeven2024point}, using techniques that appropriately balance speed and performance. \trace's dependency graph-based architecture facilitates study of multimodal fusion \cite{khebour2025feature}, and adaptation to other situations and tasks by easily substituting or adding models and features. For example, modules for posture classification can be introduced to model social dynamics and level of individual task engagement \cite{moulder2022assessing,adams2022marginality}. Additionally, \trace's codebase has already been leveraged in ongoing work as a flexible platform that can support multiple different research and demonstration efforts. One such example is \citet{palmer2024speechenoughinterpretingnonverbal}, which uses the underlying \trace~platform to track nonverbal indicators of group engagement, such as joint visual attention and posture. 
\trace~will be of use to researchers in dialogue studies and collaborative problem solving, and can be used in building AI systems that mediate collaboration, such as by inserting probing questions \cite{karadzhov2023delidata,nath-etal-2024-thoughts} at key moments.


Adaptation of real-time common ground tracking with \trace~to other collaborative task scenarios is straightforward. Many modules use off-the-shelf processors like Whisper ASR, openSMILE, and MediaPipe \cite{lugaresi2019mediapipe}. Models for epistemic classification and propositional extraction can be trained on annotated data. Propositions for a new task can be deterministically enumerated following \citet{venkatesha2024propositional}, and the faster but less accurate cosine similarity method requires no new model. Our gesture recognition models can be reused as long as participants' are positioned similarly to the Weights Task. There is a practical limit of $\sim$5 bodies within the camera FOV.

Future improvements to \trace~as used in the Weights Task include also tracking the individual beliefs about the task not shared by the group, moving away from specialized depth cameras through RGB versions of modules like gesture recognition, and improving epistemic move classification through richer representation of modalities like gesture and facial expression. Further improvements to and use cases for \trace~include deploying it in less-constrained, more flexible tasks where conversations may be more ambiguous or more diverse, range in different directions, and cover a wider potential space of propositions. We are currently working on expanding \trace's usage into such tasks, such as collaborative construction and annotation of non-verbal indicators in general collaborative settings. Additionally, we continue to improve \trace's flexibility and codebase organization to allow it to accommodate new models and custom technologies, further permitting researchers to deploy individualized solutions for each modality and scenario of interest.  

\vspace*{-2mm}
\section*{Acknowledgments}
\vspace*{-2mm}

This material is based in part upon work supported by Other Transaction award HR00112490377 from the U.S. Defense Advanced Research Projects Agency (DARPA) Friction for Accountability in Conversational Transactions (FACT) program, and the National Science Foundation (NSF) under subcontracts to Colorado State University and Brandeis University on award DRL 2019805 (Institute for Student-AI Teaming). Approved for public release, distribution unlimited. Views expressed herein do not reflect the policy or position of the National Science Foundation, the Department of Defense, or the U.S. Government. We would also like to thank the anonymous reviewers whose feedback helped improve the final copy of this manuscript. All errors are the responsibility of the authors.

\vspace*{-2mm}
\section*{Ethical Statement}
\vspace*{-2mm}

Multimodal processing entails modeling people's speech and gesture patterns, body language, facial expression, etc., and raises questions about such technologies being used for tracking and surveillance. For example, modeling how individuals collaborate also involves at least tacitly modeling their linguistic and reasoning patterns, which may be sensitive. The WTD used for training the core modules---common ground tracking, pointing, object detection, etc.---is publicly-available anonymized data that was collected under protocols reviewed by institutional review boards for ethical research, and were conducted with subjects who consented to the release of the data. However, collaboration modeling technology should be treated cautiously when it comes to ingesting multiple modal channels from specific people.

\bibliography{anthology,custom}

\appendix

\section{Technical Specifications of Individual Modules}
\label{app:modules}
\vspace*{-2mm}

\paragraph{Automatic Speech Recognition}
We run Whisper at {\tt float16} precision.

\paragraph{Object Detection}
FasterRCNN was initialized with the default ResNet-50-FPN weights from TorchVision and trained 10 for epochs with batch size 32, input size 3$\times$416$\times$416, SGD with learning rate $1e-3$, momentum $9e-1$, and weight decay $5e-4$.

\paragraph{Gesture Recogntion}
We use hand features extracted from depth video using MediaPipe~\cite{lugaresi2019mediapipe} as inputs to the gesture recognizer. For the ``near'' and ``far'' radii for the pointing frustum of \citet{vanderhoeven2024point}, we use 40mm and 70mm, respectively.

\paragraph{Gaze Detection}
In the absence of eye tracking, we use direction of participants' noses as a proxy for gaze direction. This is extracted from the body rigs recognized using the Azure Kinect SDK, which consist of directed acyclic graphs containing 32 ``joints.'' We  average both ear joints, resulting in a point roughly behind the nose, and gaze direction is calculated using the vector between this point and the nose joint. Like \citet{vanderhoeven2024point}, we extend this vector out into 3D space to see which objects participants' gazes are landing on. Averaging the locations of both eyes and the nose resulted in a stable prediction that matched the direction of the participant's gaze. Because participants are always looking at something even if they aren't focusing on it (unlike intentional deixis), objects are not considered ``selected'' by gaze, but gaze may be used as a secondary feature.

\paragraph{Epistemic Move Classifier}
The epistemic move classifier we used is slightly modified from the one appearing in \citet{khebour-etal-2024-common-ground}. openSMILE features were normalized using min-max scaling. SMOTE~\cite{chawla2002smote} was used for oversampling the data, as in \citet{khebour-etal-2024-common-ground}, but when used for discrete features can create invalid data. For example, collaborative problem solving (CPS) facets~\cite{sun2020towards} used in training the model are supposed to be binary values, but SMOTE can output continuous values, so synthetic values are rounded to the nearest binary value. Finally, we also include a ReLU layer after the first linear layer for each modality. See \citet{khebour-etal-2024-common-ground} for other model specifications, that remain unchanged.

\paragraph{Propositional Extractor}
The propositional extractor from \citet{venkatesha2024propositional} proved to be limited by the sparsity of propositions actually expressed in the task (a total of 128) compared to the total number of propositions that could be expressed in the domain (total of 5,005). For example, while $yellow + purple + green > red$ is a possible proposition according to the combinatorics of the objects, it is extremely unlikely to ever actually be expressed during task performance (because the combination of yellow, purple, and green blocks so obviously outweigh the red block that groups never even need to try this). Meanwhile $green + purple = yellow$ is much more likely but may be sparsely represented in actual data (only occurring once in a group if at all). Therefore we improved cross-encoder performance using data augmentation. We prompted GPT-4 through its API to create 10 utterances that expressed each of the 128 propositions that occurred in the actual data. The model was then trained on the original transcript utterances augmented with this set. The GPT system prompt is given below, which is followed by the specific proposition for which we generated supplementary corresponding utterances. The generated utterances were subsequently human-validated for correctness before model training.


The cross-encoder was trained to output a score $Score(u_i,p_j)=MLP([V_{CLS},V{u_i},V{p_j},V_{u_i} \odot V_{p_j}$]) for an utterance $u_i$ and a candidate proposition $p_j$ over the concatenated representations of the BERT {\tt [CLS]} token for the utterance-proposition sequence, the individual utterance and proposition, and their Hadamard product, using the same hyperparameters reported in \citet{venkatesha2024propositional}.

Where \citet{venkatesha2024propositional}'s heuristic pruning left more than 137 candidate propositions, cross-encoder inference became slower than performing a vector-similarity comparison against all propositions in the vocabulary. In these cases, we back off to the cosine similarity method from \citet{khebour-etal-2024-common-ground}.

\vspace*{-1mm}
\begin{tcolorbox}[title={\sc GPT System Prompt for Propositional Data Augmentation}]
\small Conversation Background: Participants are first given a balance scale to determine the weights of five colorful wooden blocks. They are told that the red block weighs 10 grams, but that they have to determine the weights of the rest of the blocks using a balance scale. \\

The possible weights of the blocks are 10, 20, 30, 40, 50. 
Propositional content in the Weights Task takes the form of a relation between a block and a weight value (e.g., $red = 10$), between two blocks (e.g., $red = blue$), or between one
block and a combination of other blocks (e.g., $red < blue + green$). The possible colors are red, blue, green, purple, yellow. The possible weights are 10, 20, 30, 40, and 50. The possible relations are $=$, $!$$=$, $<$, $>$. Generate 10 different utterances that could be expressed by a participant while solving this task that expresses the following proposition:
\end{tcolorbox}
\vspace*{-2mm}

\section{Substitution Study Design}
\label{app:substitution}
\vspace*{-2mm}

An ablation study as technically defined requires that the system experience ``graceful degradation" \cite{newell1975tutorial} when an input is removed. However in the case of multiparty dialogue, this is often not possible. Due to the nature of dialogue, any automated system will not perform at all in the absence of speech information or transcribed audio. Dense paraphrasing requires access to both gestures and objects simultaneously; viz. dense paraphrased text without either one of them is identical to the raw text (this is evident in the dependencies in Fig.~\ref{fig:diagram}). Thus a standard ablation study where one modality is left out entirely is not realistic. Therefore we frame our study as a ``substitution'' study \textit{a la} \citet{cohen1988evaluation} which shows the importance of each modality by allowing \trace~to look up veridical information about that modality from the dataset instead of removing it entirely. Thus our evaluation follows extremely long-standing best practices in the field of AI.

\vspace*{-2mm}
\section{Performance Profiling}
\label{app:profiling}
\vspace*{-2mm}

Table~\ref{tab:profiling} shows performance statistics for a single live CGT tracking session on a consumer-grade gaming laptop, lasting approximately 5 minutes and using 1 microphone and 1 Kinect, with 3 task participants.

\begin{table}[h!]
    \resizebox{.5\textwidth}{!}{
    \begin{tabular}{ll}
\toprule
 \multicolumn{2}{c}{\small Hardware Specifications} \\
\midrule
\small{\bf Processor} & \small 12\textsuperscript{th}-gen Intel\textregistered~Core\textsuperscript{TM} i7-12700H, 2.70 GHz \\
\small{\bf RAM} & \small 16 GB \\
\small{\bf GPU} & \small NVIDIA GeForce RTC 3070 Ti Laptop GPU \\
\small{\bf VRAM} & \small 8 GB \\
\midrule
  \multicolumn{2}{c}{\small Live Performance Usage Ranges} \\
\midrule
\small{\bf GPU} & \small 60.0--74.0\% \\
\small{\bf VRAM} & \small 4.5--5.0/8 GB \\
\small{\bf RAM} & \small 54.0--58.0\% \\
\small{- Python} & \small 42.0--44.0\% \\
\small{- \trace~Modules} & \small 12.0--14.0\% \\
\small{\bf CPU} & \small 14.0--20.0\% \\
\small{\bf FPS} & \small 5--6 \\
\bottomrule
	\end{tabular}
 }
\vspace*{-2mm}
	\caption{\label{tab:profiling}Sample performance profiling.}
\vspace*{-2mm}
\end{table}

When evaluating live performance, latency must be taken into account, however due to many factors this is difficult to assess consistently. For example, specific system hardware plays a critical role in latency, so latency time reported in one configuration may not be reliably reproduced in another configuration. The configuration reported in Table~\ref{tab:profiling} represents an approximate lower bound on the hardware that will support the version of \trace~reported in this paper, and so the reported frame rate of 5--6 FPS can be taken as an approximate upper bound on the level of latency induced by processing that can be considered acceptable performance for real-time common ground tracking.

\end{document}